\documentclass[sigconf]{acmart}
\DeclareUnicodeCharacter{2212}{-}

\AtBeginDocument{%
  \providecommand\BibTeX{{%
    \normalfont B\kern-0.5em{\scshape i\kern-0.25em b}\kern-0.8em\TeX}}}

\usepackage{balance}
\usepackage{multirow}
\usepackage{subfigure}
\usepackage[inline]{enumitem}
\setcopyright{acmcopyright}
\copyrightyear{2022}
\acmYear{2022}
\usepackage{algorithm2e}
\usepackage{diagbox}
\usepackage{algpseudocode}
\def\ie{\emph{i.e.}} 
\DeclareRobustCommand\onedot{\futurelet\@let@token\@onedot}
 
\def\ie{\emph{i.e.}}

\copyrightyear{2022}
\acmYear{2022}
\setcopyright{acmcopyright}\acmConference[CIKM '22]{Proceedings of the 31st ACM International Conference on Information and Knowledge Management}{October 17--21, 2022}{Atlanta, GA, USA}
\acmBooktitle{Proceedings of the 31st ACM International Conference on Information and Knowledge Management (CIKM '22), October 17--21, 2022, Atlanta, GA, USA}
\acmPrice{15.00}
\acmDOI{10.1145/3511808.3557337}
\acmISBN{978-1-4503-9236-5/22/10}



\begin{document}


\renewcommand{\algorithmicrequire}{\textbf{Input:}}
\renewcommand{\algorithmicensure}{\textbf{Output:}}

\title{GraTO: Graph Neural Network Framework Tackling Over-smoothing with Neural Architecture Search}






\author{Xinshun Feng}
\affiliation{%
  \institution{Xi'an Jiaotong University}
  \city{Xi'an}
  \state{Shaanxi}
  \country{China}}
\email{2196113508@stu.xjtu.edu.cn}

\author{Herun Wan}
\affiliation{%
  \institution{Xi'an Jiaotong University}
  \city{Xi'an}
  \state{Shaanxi}
  \country{China}}
\email{wanherun@stu.xjtu.edu.cn}

\author{Shangbin Feng}
\affiliation{%
  \institution{University of Washington}
  \city{Seattle}
  \state{WA}
  \country{USA}}
\email{shangbin@cs.washington.edu}

\author{Hongrui Wang}
\affiliation{%
  \institution{Xi'an Jiaotong University}
  \city{Xi'an}
  \state{Shaanxi}
  \country{China}}
\email{wanghongrui@stu.xjtu.edu.cn}

\author{Qinghua Zheng}
\affiliation{%
  \institution{Xi'an Jiaotong University}
  \city{Xi'an}
  \state{Shaanxi}
  \country{China}}
\email{qhzheng@mail.xjtu.edu.cn}

\author{Jun Zhou}
\affiliation{%
  \institution{Ant Group}
  \city{Xi'an}
  \state{Shaanxi}
  \country{China}}
\email{jun.zhoujun@antfin.com}

\author{Minnan Luo}
\affiliation{%
  \institution{Xi'an Jiaotong University}
  \city{Xi'an}
  \state{Shaanxi}
  \country{China}}
\email{minnluo@xjtu.edu.cn}



\renewcommand{\shortauthors}{Xinshun Feng et al.}

\begin{abstract}
Current Graph Neural Networks (GNNs) suffer from the over-smoothing problem, which results in indistinguishable node representations and low model performance with more GNN layers. Many methods have been put forward to tackle this problem in recent years. However, existing tackling over-smoothing methods emphasize model performance and neglect the over-smoothness of node representations. Additional, different approaches are applied one at a time, while there lacks an overall framework to jointly leverage multiple solutions to the over-smoothing challenge. To solve these problems, we propose GraTO, a framework based on neural architecture search to automatically search for GNNs architecture. GraTO adopts a novel loss function to facilitate striking a balance between model performance and representation smoothness. In addition to existing methods, our search space also includes DropAttribute, a novel scheme for alleviating the over-smoothing challenge, to fully leverage diverse solutions. We conduct extensive experiments on six real-world datasets to evaluate GraTo, which demonstrates that GraTo outperforms baselines in the over-smoothing metrics and achieves competitive performance in accuracy. GraTO is especially effective and robust with increasing numbers of GNN layers. Further experiments bear out the quality of node representations learned with GraTO and the effectiveness of model architecture. We make the code of GraTo available at Github (\url{https://github.com/fxsxjtu/GraTO}).
\end{abstract}



\maketitle
\begin{figure}
    \centering
    \subfigure[]{
    \includegraphics[scale=0.5]{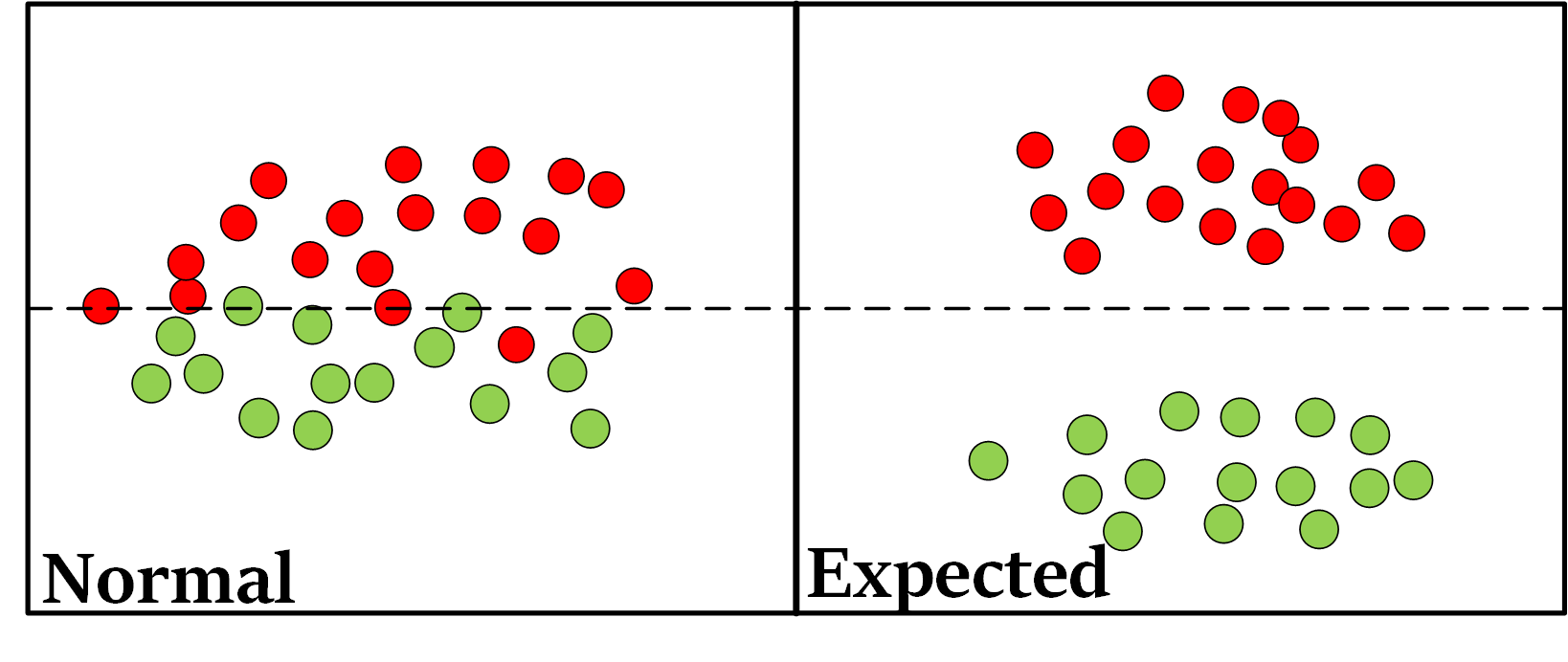}
    \label{pic:teasera}
    }
    \subfigure[]{
    \includegraphics[scale=0.7]{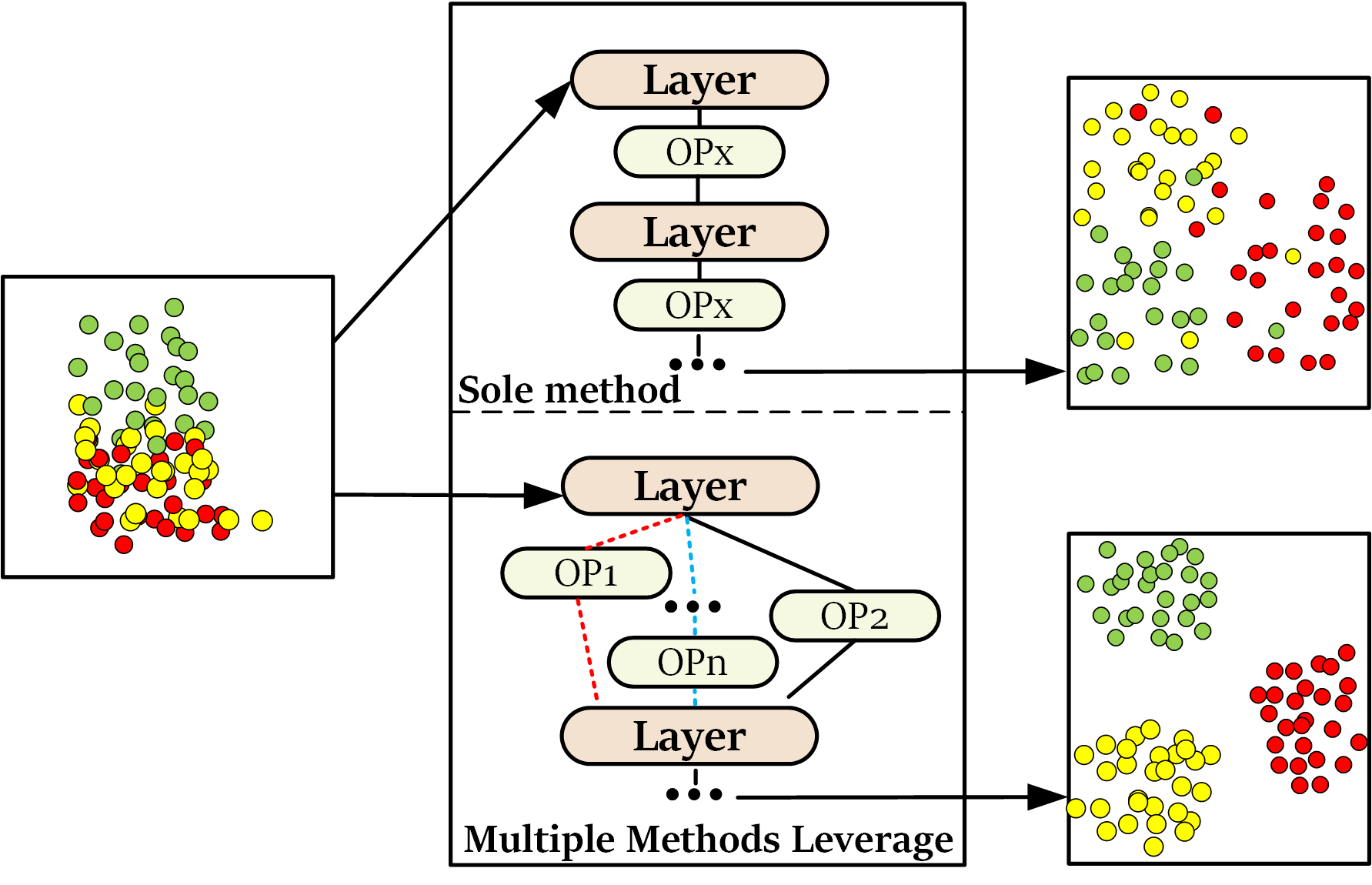}
    \label{pic:teaserb}
    }
    \caption{Illustration of drawbacks of existing tackling over-smoothing methods. (a) compares node representation between normal methods and our expectation. (b) compares model performance between sole method and diverse methods leverage.}
    \label{teaser}
\end{figure}

\section{Introduction}\label{sec:1}
Graph Neural Networks (GNNs) are a series of powerful tools for processing information on non-Euclidean graph data and achieved great success in different domains, such as social network analysis~\cite{socialnetwork1,socialnetwork2}, protein structure~\cite{protein} and recommender systems~\cite{recommend1,recommend2,recommend3}. However, the depth of the model is one of the limitations of GNNs. Many researchers want to increase the number of layers to get better performances but the model performance drops significantly in fact.
Several factors are causing the drop, such as the vanishing gradients in back-propagation, overfitting due to the increasing number of parameters, as well as the over-smoothing problem. Compared to vanishing gradients and overfitting, the over-smoothing problem is relatively important. Li et al.~\cite{li2018} first proposed that after repeatedly applying Laplacian smoothing many times, the features of the nodes in the graph would converge to similar values. It leads different classes of nodes to become indistinguishable, which seriously affects the performance of the GNN layers.

To tackle the over-smoothing problem, many approaches had been proposed in recent years. Some methods added extra modules to the model structure to tackle the problem. Zhao et al.~\cite{pairnorm} proposed a novel normalization layer that is based on an analysis of the graph convolution operator. The normalization layer ensures that the total pairwise feature distances remain constant across layers, which in turn leads to distant pairs having less similar features, preventing feature mixing across clusters. Zhou et al.~\cite{dgn} also proposed a differentiable group normalization. The group normalization can normalize nodes within the same group independently to increase their smoothness and separate node distributions among different groups. Chen et al.~\cite{gcnii} utilized identity mapping as a supplement for initial residual connection to tackle the problem. There are also some methods trying to tackle the problem by personalizing the information aggregation for each specific node. Gasteiger et al.~\cite{appnp} proposed an algorithm to separate the neural network from the propagation scheme. The algorithm balances the needs of preserving locality and leveraging the information from a large neighborhood. Liu et al.~\cite{dagnn} made the model learn node representations by adaptively incorporating information from a large receptive field. Apart from these methods, Rong et al.~\cite{dropedge} randomly removed a certain number of edges from the input graph at each training epoch to slow down the convergence of over-smoothing. Chen et al.~\cite{mad} added a specifically designed regularizer to the training objective and a new algorithm that optimizes the graph topology based on the model predictions. 

The previous methods mentioned above have two main drawbacks as shown in Figure \ref{teaser}: 
\begin{enumerate*}
\item As Figure \ref{pic:teasera} shows they focus on the model's performance but neglect the over-smoothness of node representation. 
It results in that different-label nodes remain close and the node clusters are not distinguishable.
\item As Figure \ref{pic:teaserb} shows that the existing methods only consider applying one method at a time without considering an overall framework to jointly leverage multiple methods to the over-smoothing problem. We argue that adding extra modules to the model structure or applying some operations to the model is not sufficient enough. 
There is an urgent need for a new framework to jointly leverage all methods.
\end{enumerate*}

To these issues, we propose a novel framework named GraTO (\textbf{Gra}ph Neural Network Framework \textbf{T}ackling \textbf{O}ver-smoothing). We apply neural architecture search to solve over-smoothing problem by utilizing existing GNN layers, existing methods and DropAttribute. GraTO can automatically search for the best-performance model architecture for the over-smoothing problem and shows its robustness as the layer depth increases. To find an architecture that performs well on accuracy and smoothness, we design a multiple searching objective that is proved to be useful in tackling the over-smoothing problem. In summary, we highlight our main contributions as follows:
\begin{itemize}[leftmargin=*]
    \item 
    We first propose a framework based on differentiable graph neural architecture search which takes both the architecture of the model and diverse tackling over-smoothing methods into consideration. It searches for the best block architecture by optimizing the learnable parameters and architecture parameters simultaneously to derive the best architecture.

    \item We propose a loss function to derive a model which performs well both in accuracy and smoothness. It is proved to be effective in node representation and promoting model performance. We also propose a series of novel tackling the over-smoothing methods called DropAttribute 
    to promote the model's performance.
    
    \item GraTO outperforms other baselines in smoothness metrics and achieves competitive results in accuracy on six benchmark datasets. Several related experiments are conducted to prove the effectiveness of loss function and DropAttribute. 
GraTO also learns better node representations than baseline methods.
\end{itemize}

\begin{figure*}[t]
    \centering
    \includegraphics[scale=0.7]{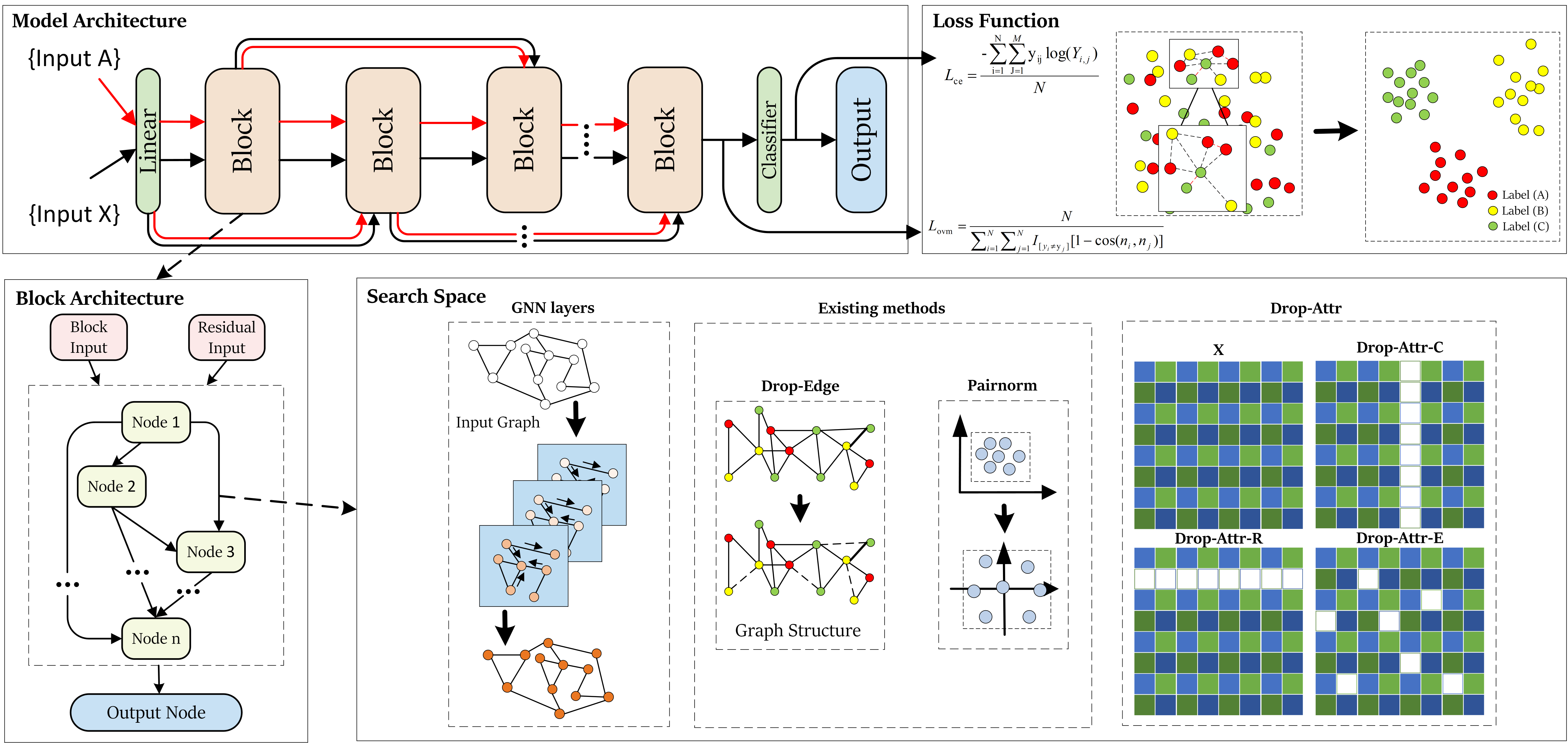}
    \caption[overview]{Overview of GraTO structure. Every block has two inputs, the direct input and residual input which is from the block right before the previous block. The block architecture briefly shows how our blocks are formed. The search space part and loss function part show the operations and loss functions we use during the search stage.}
    \label{fig:overview}
\end{figure*}

\section{Related Work}\label{relatedwork}
\subsection{Over-smoothing problem}\label{subR:1}
Taubin~\cite{first-over} first pointed out that the propagation process of GCN model is a special symmetric form of Laplacian smoothing. This smoothing process makes the representations of nodes in the same class similar, thus significantly easing the classification task. Xu et al.~\cite{xu2018} study the same problem by analyzing the connection of nodes’ influence distribution and random walk~\cite{randomwalk}. Li et al.~\cite{li2018} first formally proposed the over-smoothing problem. They defined the problem as that node features will converge to a fixed point as the network depth increases. Oono et al.~\cite{2019-over} generalized the idea in Li et al.~\cite{li2018} as convergence to a subspace rather than convergence to a fixed point. Zhao et al.~\cite{pairnorm} further defined the over-smoothing problem as repeatedly applying Laplacian smoothing too many times would drive node features to a stationary point, washing away all the information from these features. 

To measure the over-smoothing problem quantitatively, many evaluation metrics are proposed. For example, Liu et al.~\cite{dagnn} defined a similarity metric between the representations of nodes with their Euclidean distance. Chen et al.~\cite{mad} proposed Mean Average Distance (MAD) which reflects the smoothness of graph representation by calculating the mean of the average distance from nodes to other nodes. Zhou et al.~\cite{dgn} proposed group distance ratio and instance information gain. The group distance ratio is the average of pairwise representation distances between two different groups (within a group) and instance information gain is defined by how much input feature information is contained in the final representation. 

To tackle the over-smoothing problem, many methods are proposed in recent years. For example, there are many normalizations designed. Ioffe~\cite{batchnorm} proposed batch normalizations that take a step toward reducing internal covariate shift. Zhao et al.~\cite{pairnorm} proposed a novel pair normalization layer for the GNN layers to ensure that the total pairwise feature distances remain constant across layers. Zhou et al.~\cite{dgn} proposed a differentiable group normalization to softly cluster nodes and normalize each group independently.
\subsection{Neural Architecture Search}
Neural architecture search (NAS) is proposed to automatically search for the architecture which suits the searching objective. The framework of NAS can be easily divided into two types: differentiable search techniques such as DARTS~\cite{darts} and non-differentiable search techniques based on reinforcement learning~\cite{rlNAS} or evolution ~\cite{evNAS}. Liu et al.~\cite{darts} initially proposed DARTS which mainly consists of search space and search objective. The search space of DARTS is continuously relaxed by placing a mixture of candidate operations on each edge. Due to its search algorithm, its search objective is mainly decided by the loss function. But regarding the effectiveness and generalizability of gradient methods for solving non-convex architecture hyperparameter optimization problems, this method still has problems. Recently, Mills et al.~\cite{cikmnas1} proposed a novel neural architecture search called $L^{2}$ NAS. It learns to generate architecture hyper-parameters $\alpha$ in a continuous domain via an actor network based on MLP and trains the model with an effective and efficient procedure through a quantile-driven loss. However, these works are initially applied to convolutional neural networks and recurrent neural networks. Gao et al.~\cite{graphnas} made the first attempt to apply the NAS method to the graph data by utilizing existing GNN layers to build the final architecture. Zhou et al.~\cite{autognn} and Zhao et al.~\cite{snag} then further applied the NAS method to graph network data. Zhou et al.~\cite{autognn} proposed a more efficient controller which considers the key property of GNN architecture and they slightly modified the variation of representation learning capacity. Zhao et al.~\cite{snag} proposed a novel and effective search space which is more flexible than previous work. They used the reinforcement learning search algorithm in the searching stage. Khawar et al.~\cite{cikmnas2} applied NAS to click-through Rate (CTR)Prediction. They introduce NAS for learning the structure of feature interactions in neural network-based CTR prediction models. They presented a NAS search space for the CTR prediction task and a NAS algorithm (NBTree) that recursively reduces the search space.

\section{Methodology}\label{method}

\subsection{Problem Formulation}\label{subsection:problem}
Traditional GNNs typically suffer from severe over-smoothing problem as the number of layers increases. To balance the model performance and node representation smoothness, there is an urgent need to find a classification function $f:({Y},{\tilde{X}})=f(X,A)$, where $X$ and $A$ denote the input feature map and adjacency matrix, respectively; We denote the hidden representation as node representation $\tilde{X}$, and $Y$ as the predicted label matrix. The aim of this work is to find a $f$ that maximizes $\tilde{X}$ at smoothness metrics and $Y$ at accuracy metrics on the test split.

\subsection{Model Architecture}\label{modelstructure}
The overview of GraTO is presented in Figure \ref{fig:overview} that consists of model structure, block architecture, search space and loss functions. For convenience, we define the search space as $P$ in the following sections. 
Inspired by Resnet~\cite{resnet}, we introduce the residual connection to GraTO. Every block has two inputs direct input and residual input. 
We denote the last block output $\tilde{X}$ as node representation and process it with a classifier layer as the label prediction $Y$.
 
The inside architecture of the block is defined as a directed acyclic graph. It consists of an ordered sequence of $n+3$ nodes, including $n$ intermediate nodes, 2 input nodes and 1 output node. Every intermediate node has $k$ edges with the previous nodes and every edge denotes a specific operation. 
The input for node \(N_i\) is the combination of representations of feature map \(x_i\) and adjacency matrix \(a_i\). Every directed edge \((i,j)\) represents some operations $f_{i,j} \subset P$ transforming \(x_i\) and \(a_i\), \ie,
\begin{equation}
(x_{i,j,k},a_{i,j,k}) = f_i,_j^k(x_i,a_i),
\end{equation}
where \(f_i,_j^k\) denotes the $k$-th operations in \(f_i,_j\); \(x_i,_j,_k\), \(a_i,_j,_k\) denotes the result of \(f_i,_j^k\) transforming \(x_i\) and \(a_i\). We define the number of operations in $P$ as $o$ for convenience. To differentiate the importance of every operation in the GraTO, we introduce the learnable parameters \(\lambda_{i,j,k}\) (\(k=1,2,\cdots,\)$o$) and process \(\lambda_{i,j,k}\) to \(\alpha_{i,j,k}\) as the architecture parameter
\begin{equation}\label{eq:alpha}
\alpha_{i,j,k} = \frac {exp(\lambda_{i,j,k})}{\sum_{k'=1}^{o}exp(\lambda_{i,j,k'})},
\end{equation}
and then give it to every operation by
\begin{align}\label{eq:give}
    \begin{aligned}
    x_{i,j} = \sum_{k=1}^{o}\alpha_{i,j,k}\times x_{i,j,k},\ \ 
    a_{i,j} = \prod_{k=1}^{o}a_i,_j,_k.
    \end{aligned}
\end{align}
Given the weight, we can derive the node \(j\)'s output from all of its input predecessors
\begin{align}\label{eq:concentrate}
    \begin{aligned}
    x_j&= \sum_{i<j}x_{i,j},\ \ 
    a_j= \prod_{i<j}a_{i,j},
    \end{aligned}
\end{align}
where $\prod_{i,j}$ represents the element-wise multiply. 
We then define the output $x_{out}$, $a_{out}$ of the block
\begin{align}\label{eq:concentrate}
    \begin{aligned}
    x_{out}= \sum_{i=1}^{n} x_{i},\ \
    a_{out}= \prod_{i=1}^{n}a_{i},
    \end{aligned}
\end{align}
where $n$ denotes the number of intermediate nodes.

\begin{table}[t]
  \caption{All of the operations in search space.}
  \label{tb:1}
  \begin{tabular}{ccl}
    \toprule
    Operation Name&Function\\
    \midrule
    GCN&
    $\mathbf{X'} = \mathbf{\hat{D}}^{-1/2} \mathbf{\hat{A}}
    \mathbf{\hat{D}}^{-1/2} \mathbf{X} \mathbf{\Theta}$\\
    GAT & 
    $\mathbf{X'}=\alpha_{i,i}\mathbf{\Theta}\mathbf{X}_{i} +
    \sum_{j \in \mathcal{N}(i)} \alpha_{i,j}\mathbf{\Theta}\mathbf{X}_{j}$\\
    GraphSAGE&
    $\mathbf{X'}^{\prime}_i = \mathbf{W}_1 \mathbf{X}_i + \mathbf{W}_2 \cdot
    \mathrm{add}_{j \in \mathcal{N(i)}} \mathbf{X}_j$\\
    SGC & 
    $\mathbf{X'} = {\left(\mathbf{\hat{D}}^{-1/2} \mathbf{\hat{A}}
    \mathbf{\hat{D}}^{-1/2} \right)}^K \mathbf{X} \mathbf{\Theta}$\\
    AGNN & 
    $ \mathbf{X'} = \mathbf{P} \mathbf{X}$\\
    \multirow{2}*{Pairnorm}&    $\mathbf{\tilde{X_{i}}^c}=\mathbf{\tilde{X_{i}}}-\frac{1}{n}\sum_{i=1}^{n}\mathbf{\tilde{X_{i}}}$ \\
     & $\mathbf{\dot{X_{i}}}=\mathbf{s}\cdot\frac{\mathbf{\tilde{X_{i}^{c}}}}{\sqrt{\frac{1}{n}\sum_{i=1}^{n}\lVert\mathbf{\tilde{X_{i}}}^c\rVert_{2}^{2}}}$ \\
    Drop-Edge & 
    $ \mathbf{A'}=\mathrm{mask}(\mathbf{A},\mathbf{A_{(V_{p})}})$\\
    Drop-Attr-R & $ \mathbf{X'}=\mathrm{ mask}(\mathbf{X},\mathbf{X_{row(i)}})$\\
    Drop-Attr-C & $ \mathbf{X'}=\mathrm{ mask}(\mathbf{X},\mathbf{X_{column(j)}})$\\
    Drop-Attr-E & $ \mathbf{X'}=\mathrm{ mask}(\mathbf{X},\mathbf{X_{(V_{p}})})$ \\

  \bottomrule
\end{tabular}
\end{table}
\subsection{Search Space}\label{subsection:searchspace}
In this section, we introduce the operations in $P$. We divide them into three groups, according to their function and characteristics.

\subsubsection{Normal GNN Layers}\label{ss:1}
To aggregate the graph's information, we choose several widely used GNN layers as our basic layers.
\begin{itemize}[leftmargin=*]
\item\textbf{GCN~\cite{GCN}:}
 ${X}^{\prime} = {\hat{D}}^{-1/2} {\hat{A}}{\hat{D}}^{-1/2}{X}$ with ${\hat{A}} = {A} +{I}$.
\item\textbf{GAT~\cite{GAT}:}
 ${X}^{\prime}_i = \alpha_{i,i}{X}_{i} +
\sum_{j \in \mathcal{N}(i)} \alpha_{i,j}{X}_{j}$,
with attention coefficients 
 $\alpha_{i,j} =
\frac{
\exp\left(\mathrm{LeakyReLU}\left(\mathbf{a}^{\top}
[\mathbf{x}_i \, \Vert \, \mathbf{x}_j]
\right)\right)}
{\sum_{k \in \mathcal{N}(i) \cup \{ i \}}
\exp\left(\mathrm{LeakyReLU}\left(\mathbf{a}^{\top}
[\mathbf{x}_i \, \Vert \, \mathbf{x}_k]
\right)\right)}$.
\item\textbf{GraphSAGE~\cite{GraphSAGE}:}
 ${X}^{\prime}_i = {W}_1 {X}_i + {W}_2 \cdot
\mathrm{add}_{j \in {N(i)}} \mathbf{X}_j$,
where $W$ is the learnable parameter and the $add$ denotes the add aggregator.
\item\textbf{AGNN~\cite{AGNN}:}
 ${X}^{\prime} = {P}{X}$,
with propagation matrix $P=[P_{i,j}]$ and trainable paramete $\beta$,  
$P_{i,j} = \frac{\exp( \beta \cdot \cos({X}_i, {X}_j))}
{\sum_{k \in \mathcal{N(i)}\cup \{ i \}} \exp( \beta \cdot
\cos({X}_i, {X}_k))}$.
\item\textbf{SGC~\cite{SGC}:}
 ${X}^{\prime} = {\left({\hat{D}}^{-1/2} {\hat{A}}
{\hat{D}}^{-1/2} \right)}^K {X}$,
where $K$ is the number of hops.
\end{itemize}
\subsubsection{Existing Tackle Over-smoothing Operations}\label{ss:2}
To solve the over-smoothing problem, we utilize two existing tackling over-smoothing operations, Pairnorm~\cite{pairnorm} and Drop-Edge~\cite{dropedge}.
\begin{itemize}[leftmargin=*]
\item \textbf{Pairnorm}:
Pairnorm is used as a specific normalization layer at the output stage, and in summary, the Pairnorm can be written as a two-step procedure
\begin{align}\label{eq:concentrate}
    \begin{aligned}
    \tilde{X_{i}}^c=\tilde{X_{i}}-\frac{1}{n}\sum_{i=1}^{n}\tilde{X_{i}}, \ \
    \dot{X_{i}}=s\cdot\frac{\tilde{X_{i}^{c}}}{\sqrt{\frac{1}{n}\sum_{i=1}^{n}\lVert\tilde{X_{i}}^c\rVert_{2}^{2}}},
    \end{aligned}
\end{align}
where $\lVert\tilde{X_{i}}\rVert_{D}^{2}=X_{i}^{T}DX_{i}$; $\tilde{X}$ and $\dot{X}$ denotes the input  and output, respectively; hyperparameter $s$ determines constant total pairwise
squared distance $C$ in the Equation. 
\end{itemize}

The pairnorm is proved to have a great influence on tackling over-smoothing problem in related experiments. We apply this normalization layer after every GNN layer output. 
\begin{itemize}[leftmargin=*]
\item \textbf{Drop-Edge}:
As the related experiments in~\cite{dropedge} shows, Drop-Edge reveals higher distance and slower convergent speed, which can be seen as tackling over-smoothing problem. The equation of Drop-Edge can be written as:
 $A' =\mathrm{mask}(A,A_{(V_{p})})$.
We define $A_{(V_{p})}$ as the matrix which only contains 0 or 1 and its $V_{p}$ part of elements have been set to 0 and the rest are 1, where $V$ is the total number of non-zero elements and $p$ is the dropping rate. The mask operation refers to the element-wise multiply. Compared to Drop-Edge in the original paper, we don't perform the re-normalization trick on the derived adjacency matrix.
\end{itemize}
\subsubsection{DropAttribute}\label{ss:3}
Inspired by the Drop-Edge, we propose a series of operations to tackle over-smoothing problem named as DropAttribute. As a matter of convenience, we define the DropAttribute as Drop-Attr in the following sections. The series of operations contain: 
\begin{itemize}[leftmargin=*]
\item \textbf{Drop-attr-R:}
 $X'=\mathrm{mask}(X,X_{row(i)})$, where $X_{row(i)}$ denotes a matrix full of ones whose $i$-th row is set to zeros.

\item \textbf{Drop-attr-C:}
$X'= \mathrm{mask}(X,X_{column(j)})$, where $X_{column(j)}$ denotes a matrix full of ones whose $j$-th row is set to zeros.

\item \textbf{Drop-attr-E:}
$X'= \mathrm{mask}(X,X_{(V_{p})})$,
where $X_{(V_{p})}$ only contains 0 or 1 and its $V_{p}$ part of elements have been set to 0 while the rest are 1, $V$ is the total number of non-zero elements and $p$ is the dropping rate.
\end{itemize}
These Drop-attr operations tackle the over-smoothing problem by randomly removing several nodes' values on the feature map. We conduct extensive experiments to further prove their effectiveness in the experiment section. We show all the operations in Table \ref{tb:1}.

\subsection{Train and Inference}\label{subsection:searchobjective}
\subsubsection{Loss Function}
In this section, we show the loss function used and optimize objective in the training stage. As the Section \ref{subsection:problem} implies, the output of our model can be defined as the output label prediction $\mathbf{Y}$ and node representation $\mathbf{\tilde{X}}$ at the last hidden layer. We define $y$ as ground-truth label and the whole loss function as
\begin{equation}\label{eq:loss}
    L= \lambda  L_{ovm}(\mathbf{\tilde{X}},y)+L_{ce}(Y,y),
\end{equation}
where the $\lambda$ is the coefficient of $L_{ovm}$ to adjust the whole loss function. Inspired by the sampling loss function proposed by Chen et al.~\cite{loss}, we define the pairs of nodes which are in the same labels as "positive" pairs of nodes, and other pairs are "negative" pairs of nodes. we randomly sample $N_{s}$ pairs of nodes $n_i$, $n_j$ in the $\mathbf{\tilde{X}}$. To punish the "negative" pairs, we propose a novel loss function $L_{ovm}$. Let \(cos(x_1,x_2)=\frac{x_1\cdot x_2}{\lVert x_1\rVert\cdot\lVert x_2 \rVert}\) denotes the diversity between the node \(x_1\) and \(x_2\). Then the loss function $L_{ovm}$ is defined as:
\begin{equation}
L_{ovm}=\frac{N_{s}}{\sum_{i=1}^{N}\sum_{j=1}^{N}\mathbb{I}_{[y_i\neq y_j]}[1-cos(n_i,n_j)]},
\end{equation} where \(\mathbb{I}_{[y_i\neq y_j]}\in \{ 0,1\} \) is an indicator function evaluating to 1 iff node \( i\) and node \(j\) get the different labels in labels \(y\). The loss function $L_{ce}$ can be written as
\begin{equation}
L_{ce}(Y, y) = -\frac{ \sum_{i=1}^{N}\sum_{j=1}^{M} y_{ij}\log(Y_{ij})}{N},
\end{equation}
where $M$ is the number of classes and $N$ denotes the number of nodes, $y_{ij}$ denotes an indicator evaluating to 1 iff node $i$'s prediction equals to the label $j$ or 0 in other situations and $Y_{ij}$ denotes the probability of node $i$ in label $j$.

\SetKwComment{Comment}{/* }{ */}
\RestyleAlgo{ruled}
\begin{algorithm}[t]
\caption{Search Process in GraTO}
\label{algo}
\KwData{feature map $X$, adjacency matrix $A$}
\KwResult{block architecture}
Initialize parameters $\omega$ and learnable parameters $\lambda$\;
Initialize block list $\mathbb{L}$\;
Search start\;
\While{$\omega$ not converged}{
    compute $\alpha_{i,j,k} \gets \lambda_{i,j,k}$ from Equation (\ref{eq:alpha});

    $x_{i,j,k} \gets \alpha_{i,j,k}$ following Equation (\ref{eq:give});

    Derive ${\tilde{X}}$ as last hidden representation and $Y$ as output following Equation (\ref{eq:concentrate});

    Update architecture $\alpha$ by descending $\nabla_{\alpha} L_{val}(\omega ^{*}(\alpha),\alpha)$;
    
    Update weights $\omega$ by descending $\nabla_{\alpha} L_{train}(\omega,\alpha)$;

    Derive the model architecture by retaining the top-k strongest operation determined by Equation (\ref{eq:alpha});

    Test the block on the validation split;
    
    \eIf {\rm the model outperforms previous in $\mathbb{L}$} {$\mathbb{L}$.update(model);}{continue;}
  }
Choose the last block architecture in $\mathbb{L}$ as the final block, then train and test the block;
\end{algorithm}

\subsubsection{Optimize Objective }
To search for a block that performs well on accuracy and smoothness, we define the \(\omega\) as the learnable parameter of the model, $\alpha$ as the model architecture parameter. \(L_{train}\) and \(L_{val}\) refer to the loss in the train section and validation section. Following the framework of DARTS~\cite{darts}, the objective of searching is to solve a bi-level optimization problem, \ie{
\begin{equation}\label{appro}
\begin{aligned}
\min_\alpha\  &L_{val}(w^{*}(\alpha),\alpha)\\
s.t.\ &\omega^*(\alpha)=argmin_\omega L_{train}(\omega,\alpha).\\
\end{aligned}
\end{equation}}
To make the optimization more smooth and fast, we follow the DARTS to approximate the Equation (\ref{appro}) as
\begin{equation}
\nabla_{\alpha} L_{val}(\omega',\alpha) -\frac{[\nabla_{\alpha}L_{train}(\omega^+,\alpha)-\nabla_{\alpha}L_{train}(\omega^-,\alpha)]}{2},
\end{equation}
where the \(\omega^{\pm}=\omega\pm\epsilon\nabla_{\omega^{'}}L_{val}(\omega^{'},\alpha)\).
After applying this approximation, the complexity of the optimization reduces from \(O(|\alpha||\omega|)\) to \(O(|\alpha|+|\omega|)\).
\subsubsection{Model Derivation}
\begin{figure}[t]
\centering
\includegraphics[width=0.8\linewidth]{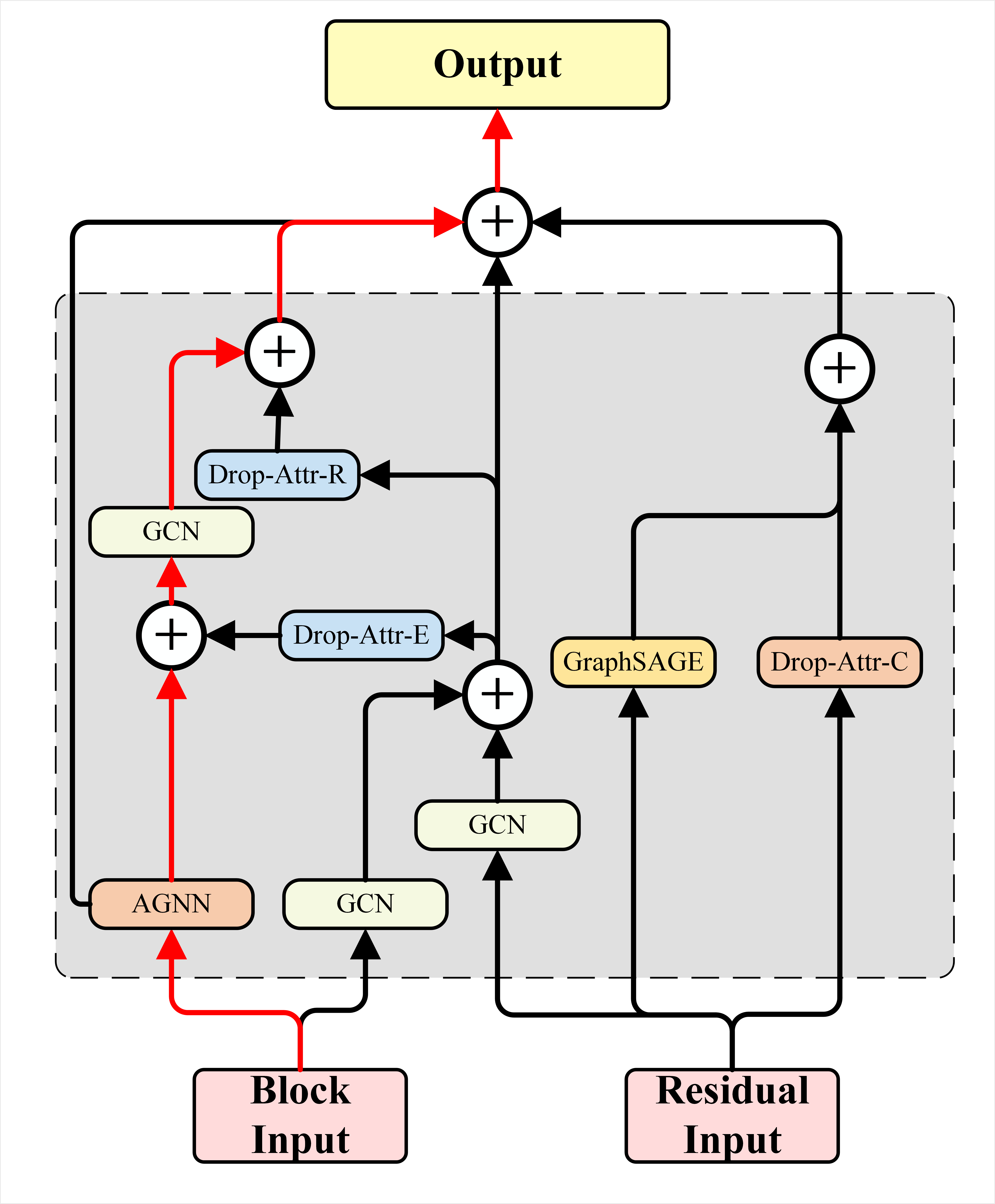}
\caption[Block Architecture found in Pubmed]{Illustration of a block containing corresponding operations. The plus symbol in the dark refers to the intermediate node, and the outside one refers to the output node. The red line represents the longest GNN layer line. We denote the red line as the longest subchain in a block.}

\label{fig:structure}
\end{figure}

During the searching stage, to form the block containing nodes in the discrete architecture, we retain the top-k strongest operations among all candidate operations collected from all the previous nodes. The strength of the operations is determined by Equation (\ref{eq:alpha}). After deriving one block architecture, we test its output $\tilde{X}$ and $Y$ on the validation split to choose the best architecture. One found architecture in Pubmed can be seen in Figure \ref{fig:structure} and the number of layers in one block is defined as the length of the longest convolutional subchain. The whole process of searching can be seen in Algorithm \ref{algo}. More detailed settings about the block are shown in Section \ref{details}.

\begin{table*}[t]
    \caption{Dataset statistical description and detailed split.}
    \label{tb:dataset}
    \begin{center}
    \begin{tabular}{lcccccccc}
    \toprule
    Dataset & Nodes & Edges & Features & Classes & Training nodes & Validation nodes & Test nodes \\
    \midrule
    Cora~\cite{cora} & 2,708 & 10,556 & 1,433 & 7 & 20 per class & 500 & 1,000 \\
    Citeseer~\cite{citeseer} & 3,327 & 9,104 & 3,703 & 6 & 20 per class & 500 & 1,000 \\
    Pubmed~\cite{pubmed} & 19,717 & 88,648 & 500 & 3 & 20 per class & 500 & 1,000  \\
    Computers~\cite{amazon} & 13,752 & 491,722 & 767 & 10 & 20 per class &  50 per class & the rest of nodes  \\
    Photo~\cite{amazon} & 7,650 & 238,162 & 745 & 8  & 20 per class &  50 per class & the rest of nodes \\
    Reddit~\cite{reddit} & 232,965 & 114,615,892 & 602 & 41 & 153,431 & 23,831 & 55,703 \\
    \bottomrule
    \end{tabular}
    \end{center}
  \end{table*}

\section{Experiment}\label{experimentall}
\subsection{Experimental Setup}

\subsubsection{Dataset}\label{subsub:1}
We use the following benchmark datasets to validate the significance of the proposed model. A detailed description of the datasets is shown in Table \ref{tb:dataset}. \begin{itemize}[leftmargin=*]
\item Cora~\cite{cora}, Citeseer~\cite{citeseer}, Pubmed~\cite{pubmed}: \label{coradata} These datasets are publications citation networks where nodes represent documents and edges represent citations. These citation networks follow the standard train/ validation/ test split following Yang et al.~\cite{semi-split}.
\item Computer~\cite{amazon}, Photo~\cite{amazon}: \label{com} These datasets are co-purchase graphs extracted from Amazon, where nodes represent products, edges represent the co-purchased relations of products and features are vectors extracted from product reviews. We split these dataset manually as 20 nodes per class in the training split, 50 nodes per class in the validation split and the rest in the test split.
\item Reddit~\cite{reddit}: It is a graph extracted from Reddit posts in September 2014. Nodes represent posts and edges represent two posts that are commented by one user. The split follows Hamilton et al.~\cite{reddit}.
\end{itemize}

\subsubsection{Baselines}\label{subsub:2}
To demonstrate the effectiveness of these methods, we choose three widely used GNN layers, GCN, GAT and GraphSAGE, as baselines. We choose several tackle over-smoothing methods considering the model characteristics.
\begin{itemize}[leftmargin=*]
\item Pairnorm~\cite{pairnorm}: It applied the first normalization layer for GNNs. It is applied in-between intermediate layers during training which has the effect of preventing over-smoothing problem. We use this method to widely used GNN layers GCN and GAT, so these baselines are written as Pairnorm-GCN and Pairnorm-GAT.
\item DGN~\cite{dgn}: It proposed differentiable group normalization to significantly alleviate over-smoothing. We apply this method to GCN and GAT, so this baselines are written as DGN-GCN and DGN-GAT.

\item ResNet~\cite{resnet}: This method introduced a deep residual learning framework to achieve a deep GNN structure. We choose to apply this method to the Pairnorm baselines which are written as Pairnorm-GAT-Res and Pairnorm-GCN-Res.
\item GCNII~\cite{gcnii}: This method proposed a simple and deep GCN model that prevents over-smoothing by initial residual connection and identity mapping. We follow the settings in the original paper, so the baseline is written as GCNII.
\item Drop-Edge~\cite{dropedge}: reduce node connection to slow down the convergence of over-smoothing. We use multi-GCN as the original paper, so the baseline is written as Dropedge-multiGCN.
\end{itemize}

\subsubsection{Evaluating Metrics}\label{subsub:3}
We choose the Mean Average Distance (MAD) value proposed by Chen et al.~\cite{mad} as an evaluating metrics for smoothness to fairly measure the smoothness of models. We choose macro F1-score and Accuracy as evaluation metrics for model performance on accuracy. To measure the overall of the model performance, we adopt an integrative ranking of F1 score, Accuracy and MAD value as an evaluating metric for model performance. To calculate the integrative ranking, we calculate the different rankings of the F1 score, Accuracy and MAD value, and make an average ranking of these rankings.
\subsubsection{Experiment details}\label{details}
In all tasks, we fix the hyper-parameters as dropping rate in Drop-Edge as 0.3, the rate in Drop-Attr-E as 0.6 and hidden dimension as 256. As for the optimizer, we apply Adam~\cite{adam} with a learning rate of 0.005 and weight decay as $1e-4$. The activation function used in every model is set as Relu. We set the number of layers of all models to 16 at Cora, Pubmed, Computers, 12 at Citeseer, Photo and 4 at Reddit due to the space limit. 
In the node representation task, we denote the last hidden representation of GNN layers as node representation. We set the dropout rate in the linear classifier layer used in the same-label node pair prediction as 0.1. Only one-layer is used for classifier.
More experiment details and codes are available at Github \footnote{\url{https://github.com/fxsxjtu/GraTO}}.

\begin{table*}[t]
\footnotesize	
	\caption{Accuracy, F1-score, MAD value and integrative ranking on the node classification task. We separate the tackling over-smoothing into three groups according to the focus: general, smoothness and accuracy and we compare all the baselines on six datasets. The "−" indicates that some baselines' training faces the out-of-memory
    problem.} 
	
	\centering
	\label{tb:wholetable}
	\begin{tabular}[width=\linewidth]{l|l|c|ccc|c|ccc|c|ccc}
	\toprule
	Focus on &\multirow{2}*{\diagbox[innerwidth=2.5cm]{Method}{Dataset}} & \multicolumn{4}{c|}{Cora}   & \multicolumn{4}{c|}{Citeseer} 
	& \multicolumn{4}{c}{Pubmed}  \\
	 class &   &Rank 
	 &Acc &F1 & MAD &Rank  &Acc &F1 & MAD &Rank  &Acc &F1 & MAD\\
	\midrule
    \multirow{3}*{General} 
     & GCN~\cite{GCN}      & 9 & 59.3 & 59.3 & 31.4 & 
			 11 & 50.8& 47.4 & 28.3& 
			 12 & 50.9 & 49.3 & 20.8\\
	 & GAT ~\cite{GAT}    & 11 & 54.6 & 54.7 & 35.9 & 
			8 & 57.2& 54.6 & 22.9& 
			9 & 69.6 & 67.2 & 33.3\\     

	 & GraphSAGE~\cite{GraphSAGE} 
	         & 11 & 57.2 & 59.0 &  26.0& 
			 12 &44.5& 43.6 & 36.9& 
			 11 &63.4 & 57.8 & 40.7\\ 
    \cline{1-2}
    \multirow{6}*{Smoothness} & 
    Pairnorm-GCN~\cite{pairnorm} & 5 & 71.6 & 69.9 &  77.9& 
			 10 & 48.3& 46.9 & 69.2& 
			 10 & 64.4 & 60.2 & 62.7\\
	 & Pairnorm-GCN-Res~\cite{pairnorm} & 4 & 72.0 & 72.6 &  73.6& 
			 4 & 57.4& 54.3 & 67.9& 
			 8 & 69.2 & 67.2 & 65.4\\ 
	 & Pairnorm-GAT~\cite{pairnorm} & 5 & 66.7 & 66.3 &  \textbf{87.1}& 
			 8 & 49.5& 47.3 & 72.5& 
			 6 & 68.3 & 67.2 & 67.3\\
    & Pairnorm-GAT-Res~\cite{pairnorm} & 3 & 74.6 & 73.2 &  71.8& 
			 3 & 59.5 & 54.9 & 69.7& 
			 3 & 74.3 & 72.9 & 67.8\\
	& DGN-GAT~\cite{dgn}  & 10 &  54.3 & 56.9 & 57.2& 
			 8 & 52.9 & 49.9 & 55.1& 
			 5 & 74.3 & 73.0 & 50.7\\
	 & DGN-GCN~\cite{dgn}  & 8 &  60.5 & 59.4 & 52.6& 
			 6 & 53.6 & 51.0& 47.3& 
			 2 & 74.9 & 73.5 & 55.1\\
	\cline{1-2}
	\multirow{2}*{Accuracy} & GCNII~\cite{gcnii}    & 2 & 81.2 & \textbf{80.3} & 35.2& 
		 2 & \textbf{70.2} & \textbf{66.3} & 47.1& 
		 4 & \textbf{80.5 }& \textbf{80.2} & 24.55\\
     & Dropedge-multiGCN~\cite{dropedge} 
     & 7 &  75.0 & 74.2 & 19.9& 
		 4 &67.4 & 64.3 & 17.8& 
		 7 & 73.3 & 73.2 & 27.6\\
    \bottomrule
	Leveraged & GraTO      & \textbf{1} & \textbf{81.5} & 80.2 &85.7& 
		     \textbf{1} & 65.2 & 61.3 & \textbf{84.5}& 
			 \textbf{1} & 78.9 & 77.7 & \textbf{87.5}\\
	\bottomrule
	\end{tabular}
	
	\begin{tabular}{l|l|c|ccc|c|ccc|c|ccc}
	\toprule
	Focus on& \multirow{2}*{\diagbox[innerwidth=2.5cm]{Method}{Dataset}} &  \multicolumn{4}{c|}{Computer} & \multicolumn{4}{c|}{Photo} & \multicolumn{4}{c}{Reddit} \\
	 Class &   &Rank 
	 &Acc &F1 & MAD &Rank  &Acc &F1 & MAD &Rank &Acc &F1 & MAD\\
	\midrule
	\multirow{3}*{General} & GCN~\cite{GCN}      & 
			 11 &47.7 & 42.6 & 38.3&
			 11 &80.1& 76.9& 33.7&
			 11 &79.8& 67.7&45.6\\
	 & GAT~\cite{GAT}     & 
			10 & 37.2 & 35.5 & 9.7&
			12 & 82.9& 80.6& 36.9&
			5 & 89.9& 79.1& 61.5\\     

	 & GraphSAGE~\cite{GraphSAGE} 
	         & 
			 12 & 58.0 & 55.7 & 30.7&
			 10 & 77.0 & 72.8 & 42.2 &
			 3 &\textbf{ 95.8}& \textbf{93.5}& 72.8  \\ 
    \cline{1-2}
	\multirow{6}*{Smoothness} & 
	Pairnorm-GCN~\cite{pairnorm} & 
			 5 & 62.9 & 66.6 & 68.8&
			 6 & 79.4 & 80.2 & 78.8 &
			 8 & 79.6& 64.5& 84.0\\
	 & Pairnorm-GCN-Res~\cite{pairnorm} & 
			 4 & 68.8 & 70.2 & 71.3&
			 4 & 79.5 & 80.6 & 79.0 &
			 2 & 89.4& 82.5& 93.9\\ 
	 & Pairnorm-GAT~\cite{pairnorm} & 
			 8 & 62.3 & 58.3 & 69.6&
			 5 & 78.3 & 80.6 & 80.8 &
			 5 & 88.5& 77.6& 67.9\\
	 & Pairnorm-GAT-Res~\cite{pairnorm} & 
			 2 & 77.1 & 77.2 & 78.5&
			 3 & 79.9 & 81.0 & 81.6 &
			 7 & 88.3& 76.9& 65.9\\
	 & DGN-GAT~\cite{dgn}  & 
			 6 & 67.3 & 66.1 & 39.8 &
			 9 & 74.7 & 65.8 & 43.2&
			 8 & 92.3 & 86.7& 58.4\\
	 & DGN-GCN~\cite{dgn}  & 
			 6 & 65.1 & 67.8 & 39.1 &
			 7 & 80.6 & 77.3 & 40.9 &
			 10 & 64.2 & 47.1 & 75.5\\
	\cline{1-2}
	\multirow{2}*{Accuracy} & GCNII~\cite{gcnii}    & 
		 3 & \textbf{80.3} & \textbf{78.6} & 47.5 &
		 2 & \textbf{93.5} & \textbf{91.8} & 44.9 &
		 4 & 92.3 & 86.7 & 58.4\\
	 & Dropedge-multiGCN~\cite{dropedge} & 
			 9 & 58.4 & 61.0 & 12.1 &
			 8 & 80.6 & 80.6 &17.1 &
			 - & - & - & -\\
    \bottomrule
	Leveraged & GraTO      & 
			 \textbf{1} & 77.3 & 76.1& \textbf{85.9} &
			 \textbf{1} & 86.1 & 85.6 & \textbf{83.9} &
			 \textbf{1} & 94.1 & 90.9 & \textbf{97.7}\\
	\bottomrule
	\end{tabular}
	
\end{table*}
\subsection{Performance on Node Classification} \label{task:nodeclassification}
\subsubsection{Model performance}
Table \ref{tb:wholetable} shows the result of all the methods on all datasets and we observe that:
\begin{itemize}[leftmargin=*]
\item GraTO outperforms other baselines at smoothness metric on five datasets and achieves a competitive result on accuracy and F1-score.

\item GCNII~\cite{gcnii} and Drop-Edge~\cite{dropedge} either change the structure of the model or change the input graph structure to promote model performance in accuracy, so they achieve great performance in the accuracy but poor in the smoothness metric.

\item Pairnorm~\cite{pairnorm} and DGN~\cite{dgn} try to use the normalization layer to separate nodes from each other so they focus on the smoothness and achieve great performance in smoothness in the result but their performance of accuracy is not competitive enough. 
\item The normal GNN layers' performance is poor, which proves the over-smoothing problem has a great influence on the normal GNN layers' performance compared to other baselines. 
\item The tackling over-smoothing baselines generally outperforms normal GNN layers in the integrative rankings. It shows that smoothness has a great influence on the model's performance.
\item GraTO's integrative ranking outperforms all other models, which demonstrates taking the model performance and node representation into consideration to achieve a better model performance, and leveraging multiple solutions outperform solely applying one method at a time.
\item We visualize the architecture discovered by GraTO on Pubmed in Figure \ref{fig:structure}. In the figure, we observe that both normal GNN layers and three different tackling over-smoothing methods are utilized together in a block. This can't be done by other existing methods which either add extra modules or apply specific operations.
\end{itemize}
\begin{figure}[h] 
\includegraphics[width=0.49\linewidth]{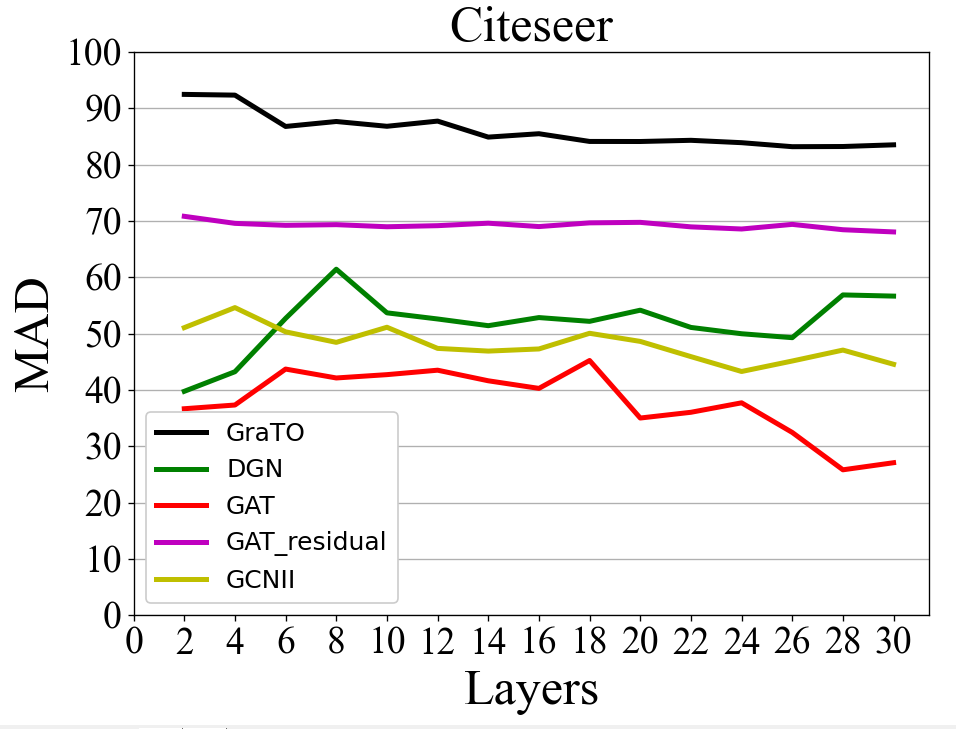}
\includegraphics[width=0.49\linewidth]{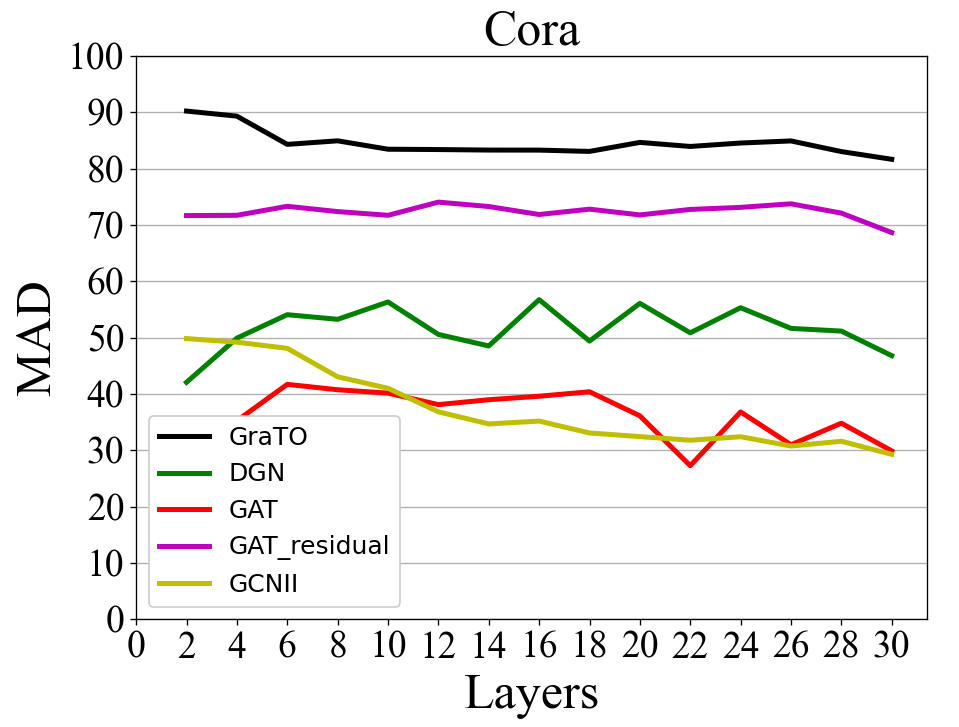}
\includegraphics[width=0.49\linewidth]{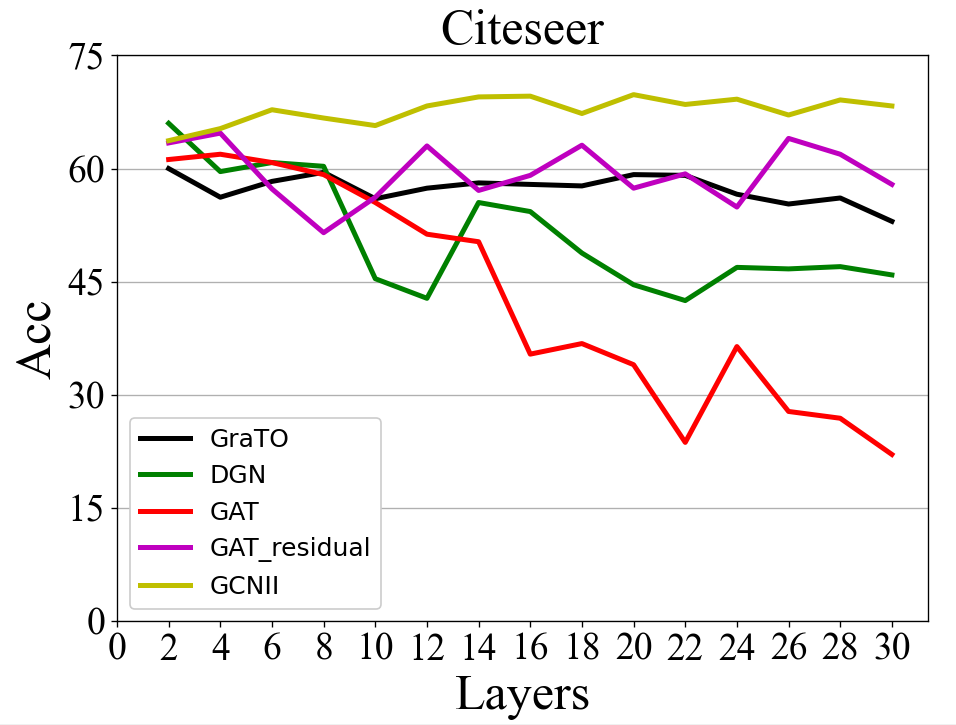}
\includegraphics[width=0.49\linewidth]{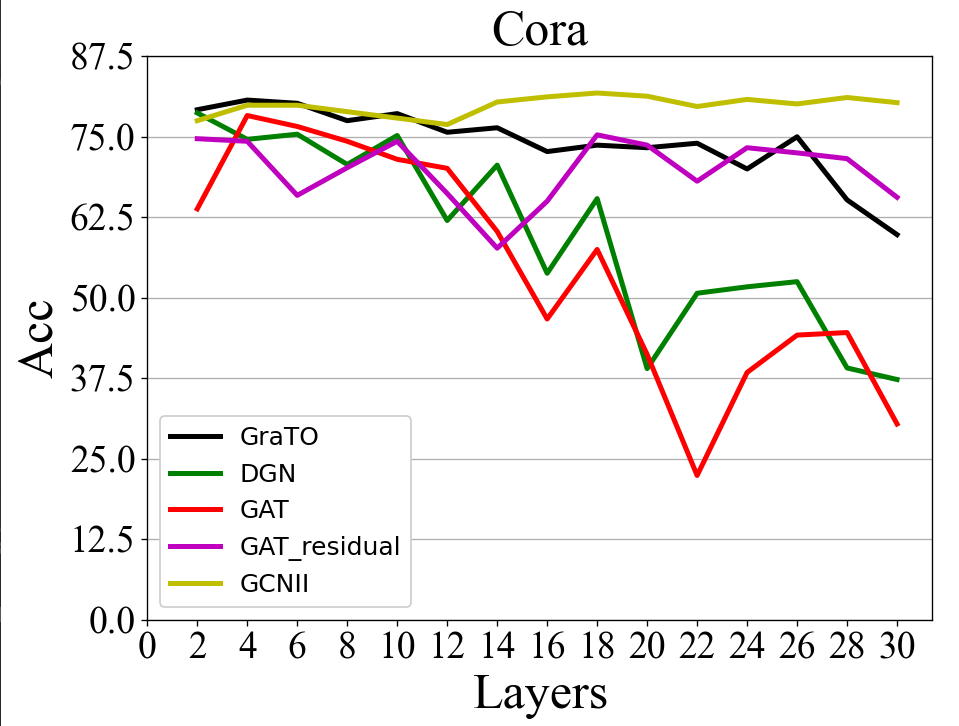}
\caption{Model performance with varying number of layers.}
\label{fig:layer}
\end{figure}
\subsubsection{Model Robustness}\label{task:2}
To prove that GraTO still performs well as the depth of layers increases, we compare the robustness of GraTO with several baselines. We calculate accuracy and MAD value while the depth of layers increases. We choose several baselines: \begin{enumerate*}
    \item Pairnorm-GAT-Res
    \item DGN-GAT
    \item GCNII
\end{enumerate*}.
We choose the GAT layer as the basic GNN layer to compare with these tackling over-smoothing methods. In Figure \ref{fig:layer} we can conclude that GraTO outperforms other baselines at smoothness metrics while the depth of layers increases. The accuracy of GraTO declines more slowly than the three of the baselines. Although GCNII~\cite{gcnii} exhibits great robustness in accuracy while the depth increases, the smoothness metric remains low which refers to its poor node representation smoothness. Pairnorm achieves relatively high performance on the smoothness metric but declines more quickly than the GraTO. All the methods mentioned above show great effectiveness in promoting normal GNN layers' performance. Jointly considering the two metrics, GraTO still achieves a great performance on the model performance and node representation smoothness. This indicates the effectiveness of considering node representation and jointly leveraging the multiple methods.

\begin{figure}[htpb] 
\includegraphics[width=\linewidth]{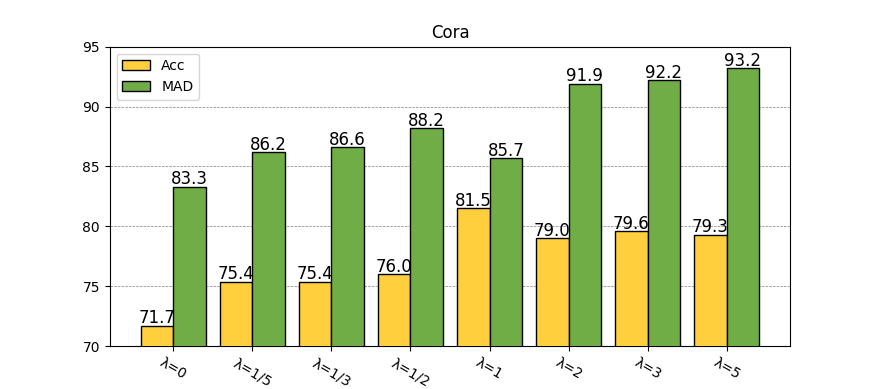}
\includegraphics[width=\linewidth]{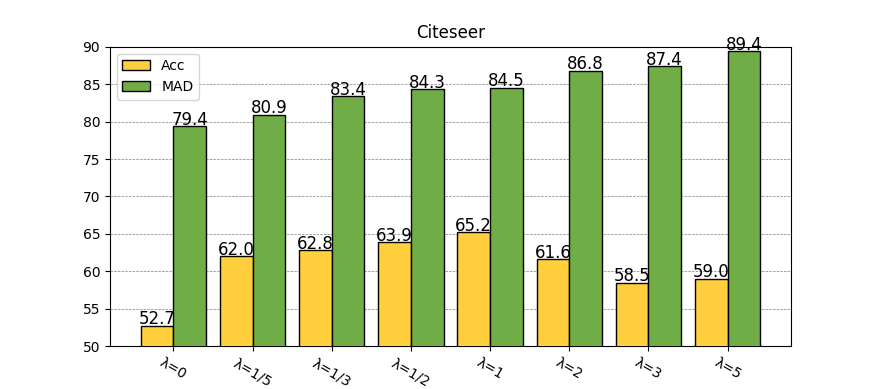}

\caption{Model performance with changing loss function weight. We change the $\lambda$ from 0 to 5 to further investigate the contributions of the loss function.}
\label{fig:loss}
\end{figure}

\subsection{Performance on Node Representation Task}\label{noderepresentation}
\subsubsection{Effectiveness of Loss Function}\label{loss}
To prove the loss function can make the node representation and model performance better, we change the $\lambda$ in Equation (\ref{eq:loss}) from 0 to 5 to investigate the model performance changes in Figure \ref{fig:loss}. We observe that with equipping our new loss function, the model outperforms the ablation one in accuracy and MAD value on both datasets. This proves that $L_{ovm}$ has a great affect on model performance and node representation smoothness. The MAD value slightly drops when $\lambda$ reaches 1 on Cora. We conjecture this is because with balancing accuracy and smoothness equally, the accuracy is promoted and smoothness is slightly dropped. The accuracy when $\lambda$ is 1 achieves the best on Cora and the accuracy when $\lambda$ is 0 achieves the worst. This indicates that considering node representation smoothness can promote the model performance in accuracy. Considering the overall performance of the model, fixing the $\lambda$ to 1 is the best option. All these results show that we should take the node representation and model performance synthetically into consideration.

\begin{figure*}[t]
\centering
    \subfigure[DGN~\cite{dgn}]{
    \includegraphics[scale=0.2]{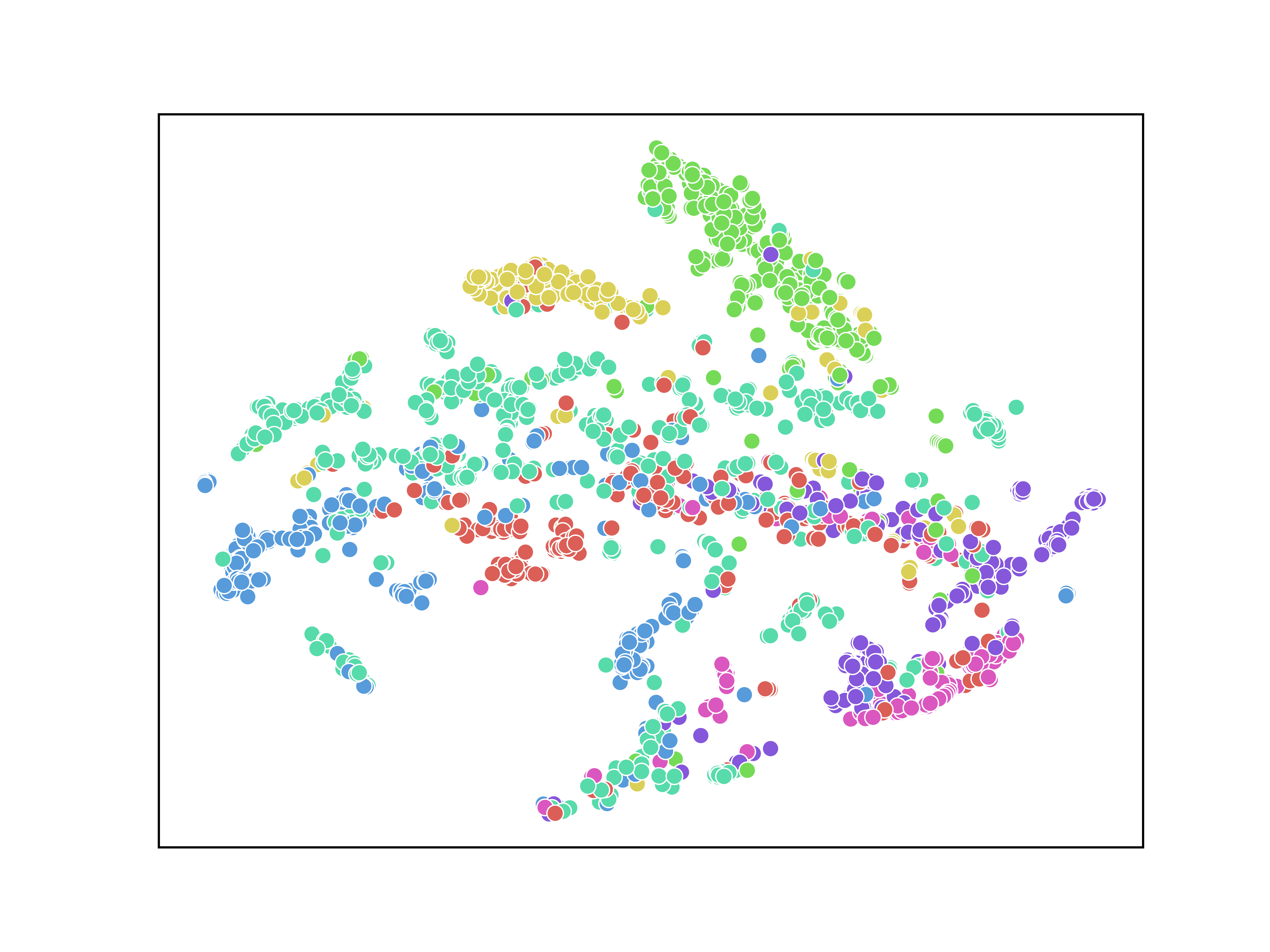}
    \label{pic:DGN}
    }
    \subfigure[GCNII~\cite{gcnii}]{
    \includegraphics[scale=0.2]{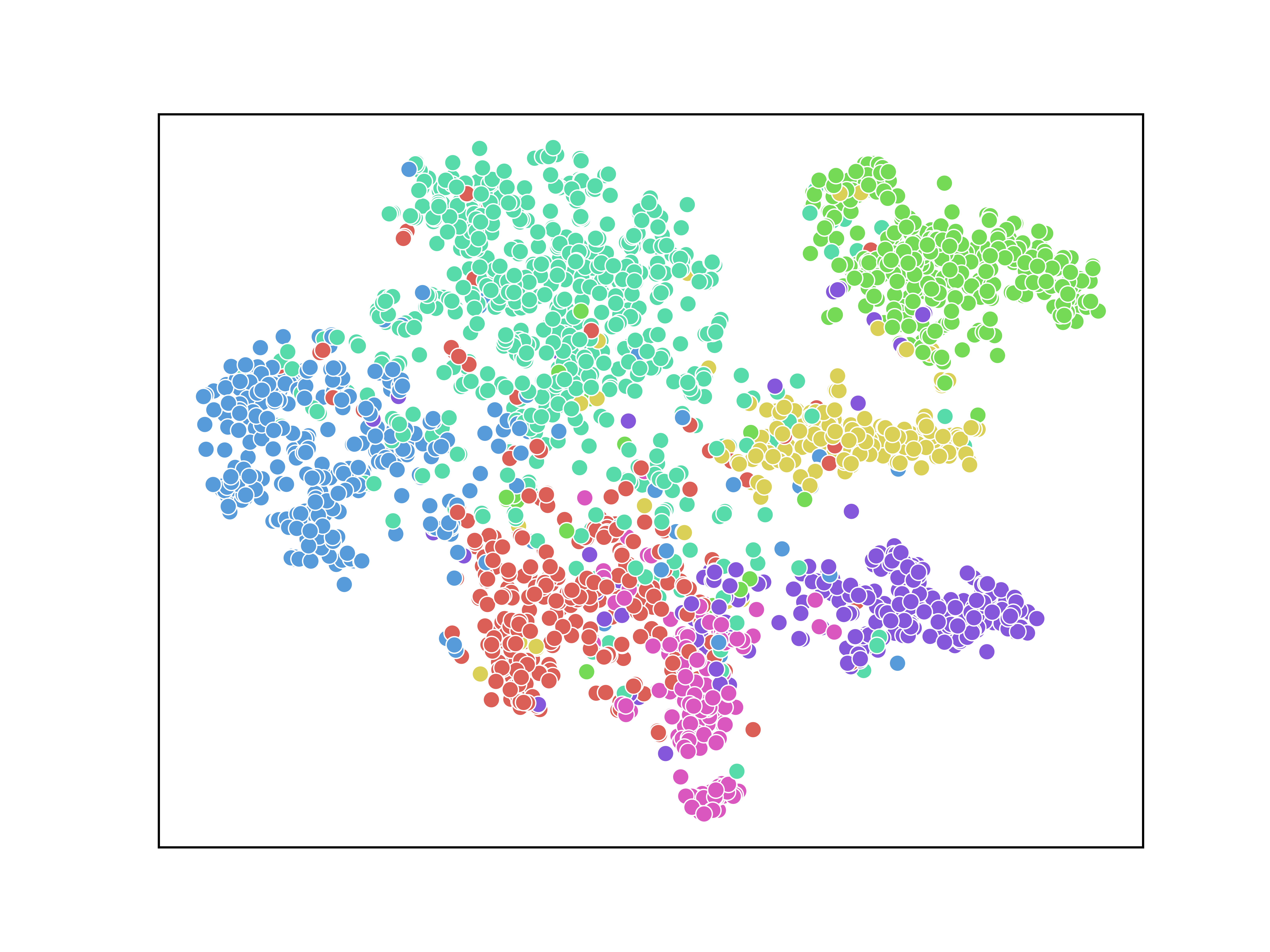}
    \label{pic:GCNII}
    }
    \subfigure[GAT~\cite{GAT}]{
    \includegraphics[scale=0.2]{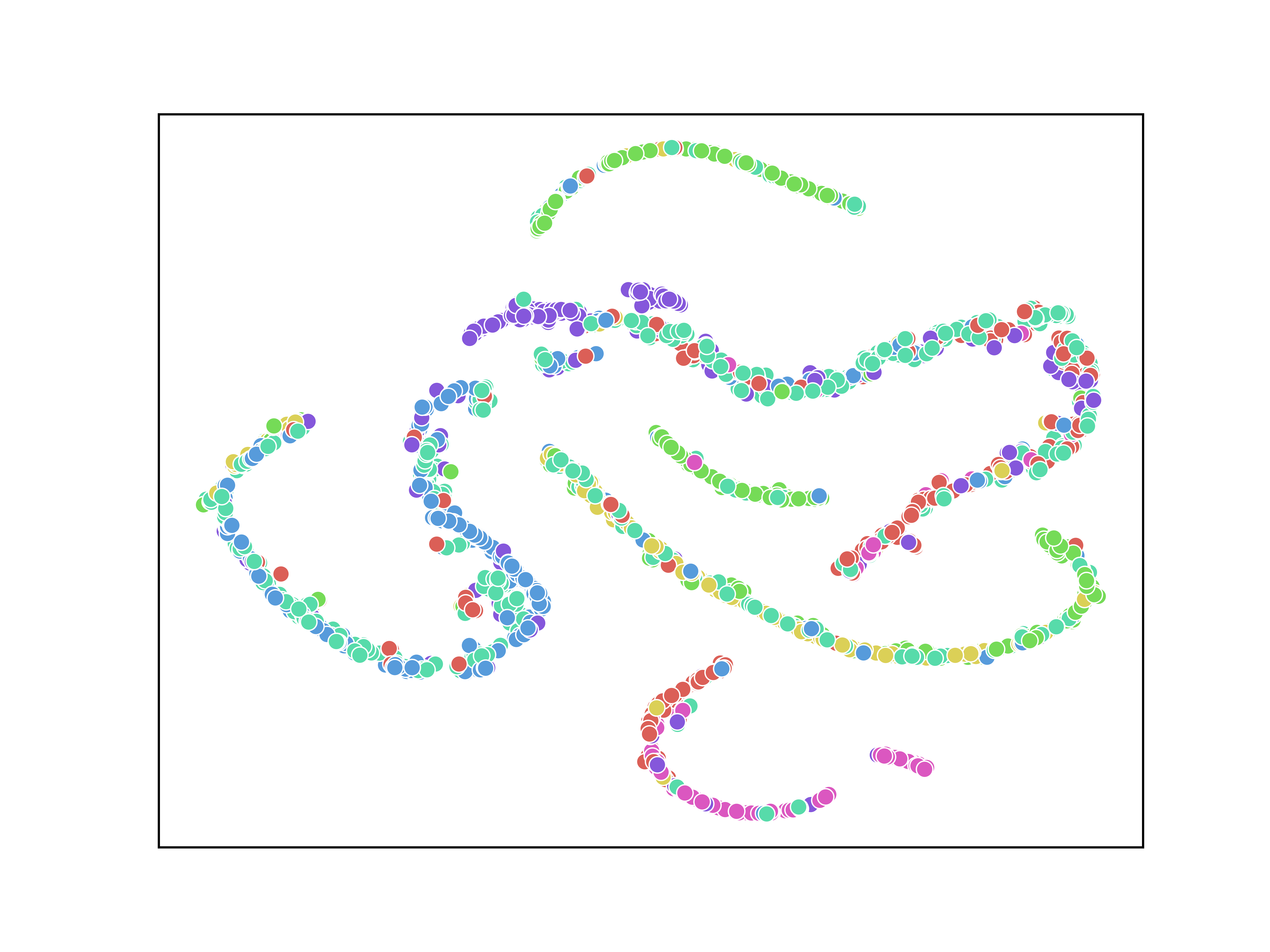}
    \label{pic:GAT}
    }
    \subfigure[Pairnorm-GAT-Res~\cite{pairnorm}]{
    \includegraphics[scale=0.2]{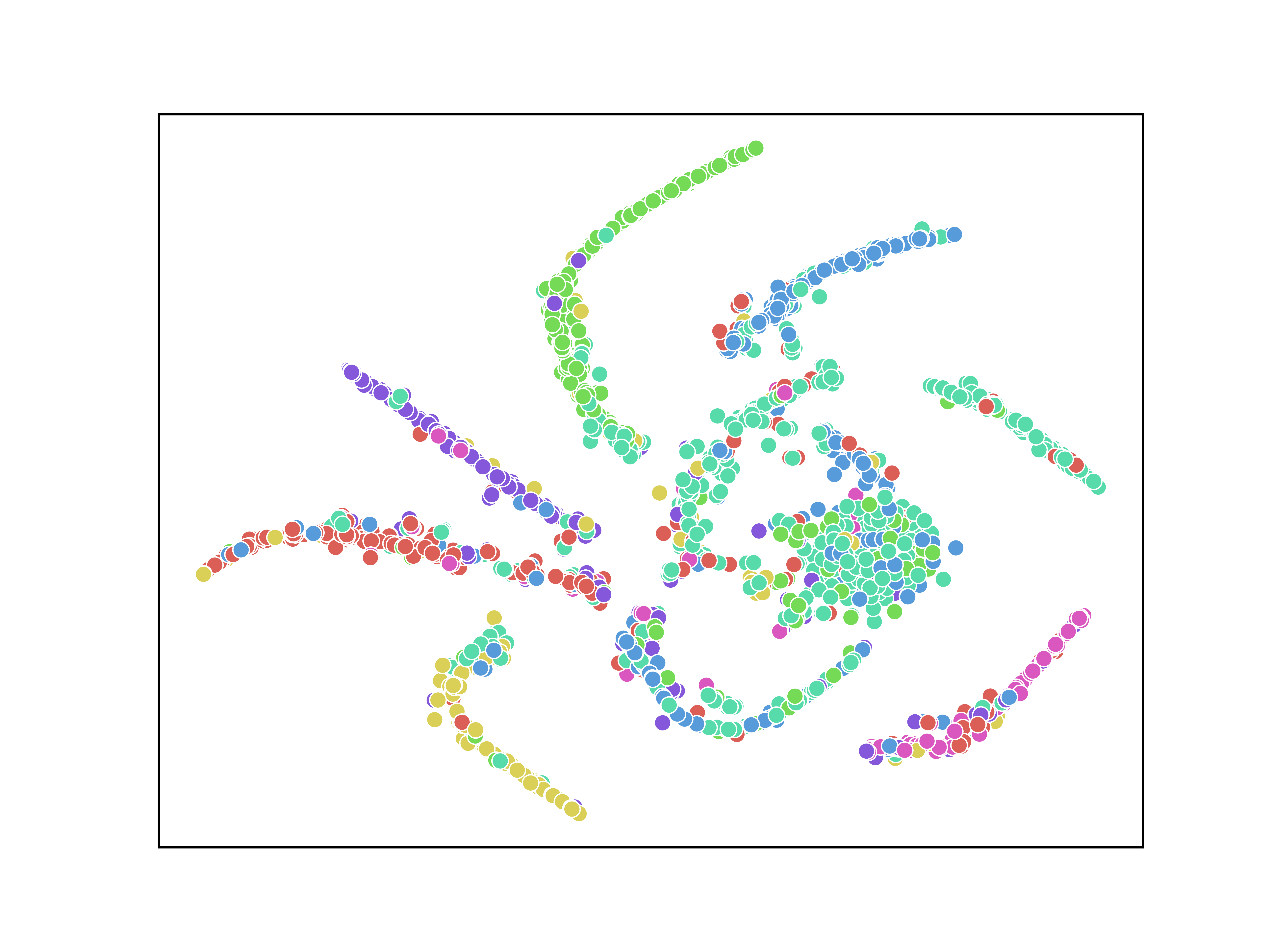}
    \label{pic:pair}
    }
    \subfigure[GraTO]{
    \includegraphics[scale=0.2]{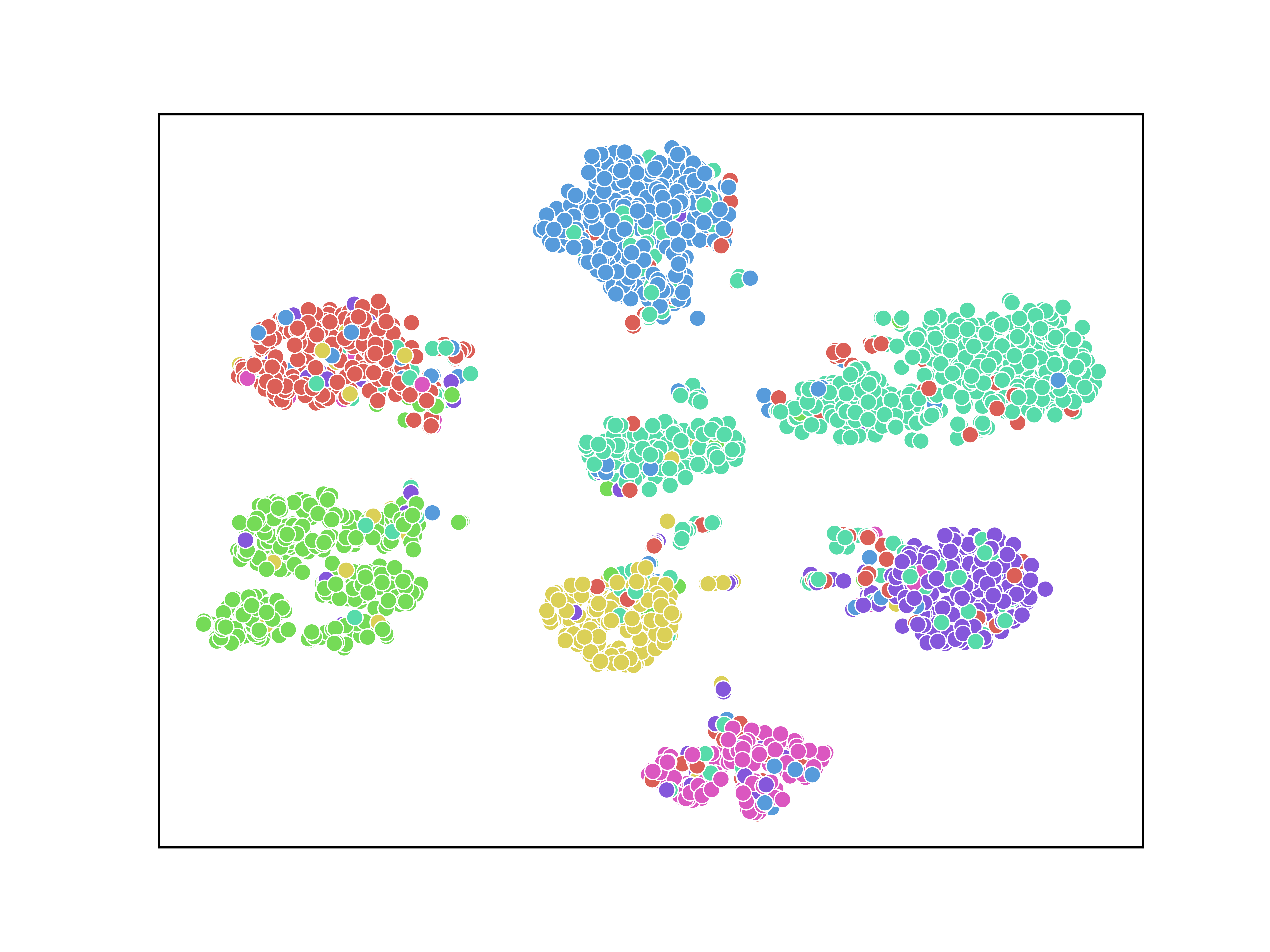}
    \label{pic:GraTO}
    }

\caption{Cluster representations of different baselines where different colors of nodes represents different labels.}
\label{fig:cluster}
\end{figure*}

\subsubsection{Clustering Coefficient Analysis}\label{cluster}
\begin{table}
  \caption{Clustering Coefficient of Different models. We compare the v-measure and $\mathrm{MAD}^{tgt}$ which is the average of different-label MAD values on the Cora and Pubmed datasets.}
  \label{tb:cluster}
  \scalebox{0.9}{
  \begin{tabular}{l|cc|cc}
  \toprule
  \multirow{2}*{\diagbox[innerwidth=3cm]{Method}{Dataset}} & \multicolumn{2}{c|}{Cora} & 
 \multicolumn{2}{c}{Pubmed}\\
   &  v-measure & $\mathrm{MAD}^{tgt}$  &  v-measure & $\mathrm{MAD}^{tgt}$ \\ 
  \midrule
    GAT~\cite{GAT}  
         & 0.3065  & 0.0999
         & 0.2240 & 0.3445 \\
    GCNII~\cite{gcnii} 
         & 0.5936 & 0.0865
         & 0.2506 & 0.1626 \\
   Pairnorm-GAT-res~\cite{pairnorm}  
         & 0.3762 & 0.2527
         & 0.0686 & 0.6093 \\
    DGN-GAT~\cite{dgn}
         & 0.3241 & 0.0068
         & 0.2512 & 0.3909 \\
    GraTO 
        & \textbf{0.6129} & \textbf{0.2681}
        & \textbf{0.2950} & \textbf{0.6196} \\
  \bottomrule
  \end{tabular}}
\end{table}
From the previous analysis, we can conclude that the over-smoothing problem results in indistinguishable nodes. To prove our node representation is good against the over-smoothing problem, we conduct the cluster analysis task. We choose the last hidden layer node representation to conduct the unsupervised learning task. We visualize the node representation of every baseline by t-SNE in Figure \ref{fig:cluster}. We demonstrate that the node representation of GraTO appears more clearly divided than the other baselines and the distance between different clusters is far. DGN and Pairnorm separate different-label nodes but the nodes in the same label appear to be separated also. GCNII divided the nodes into different groups but the cluster of nodes seems to be mixed. To measure the node representation quantitatively, we use k-means to process the last hidden representation to unsupervised clustering. We use the v-measure~\cite{vmeasure} that is a harmonic mean of homogeneity and completeness and is bounded below by 0 and above by 1. The higher homogeneity denotes that each cluster contains only members of a single class and higher completeness denotes all members of a given class are assigned to the same cluster. So the v-measure can denote the same-label nodes' aggregated degree. To measure the distance between different-label nodes' distance, we use the MAD value to measure the distance of nodes in different two labels. After calculating all different-label pairs, we take the mean of them as $MAD^{tgt}$. The results are shown in Table \ref{tb:cluster} and we can conclude that GraTO outperforms other baselines in v-measure and $MAD^{tgt}$. This indicates that our node representation achieves the long distance between different-label nodes and a high aggregated degree of same-label nodes. So as the depth of layers increases, our nodes remain strong distinguishability. As demonstrated in Chen et al.~\cite{mad}, the interaction between nodes of the same class brings useful information and the contact of nodes from other classes brings noise. As our different-label nodes remain distant and the same-label nodes remain close, nodes get more information and less noise. This indicates GraTO's high performance in smoothness.
\begin{table}[h]
  \caption{Node pair prediction result.}
  \label{tb:sample}
  \begin{tabular}{l|c|c}
  \toprule
  \multirow{2}*{\diagbox[innerwidth=3cm]{Method}{Dataset}} & \multicolumn{1}{c|}{Cora} &  \multicolumn{1}{c}{Pubmed}\\
  & AUC & AUC \\ 
  \midrule
   GAT~\cite{GAT}  
     & 0.6829  & 0.6808  \\
    GCNII~\cite{gcnii}  
      & 0.8649  & 0.7999 \\
   Pairnorm-GAT-res~\cite{pairnorm}  
      & 0.7686 & 0.5848  \\
    DGN-GAT~\cite{dgn}
     & 0.5151  & 0.7307  \\
    GraTO &
      \textbf{0.8652}  &\textbf{0.8030}  \\
  \bottomrule
  \end{tabular}
\end{table}

\subsubsection{Same-label Node Pair Predict}\label{predict}

To verify the node representation against over-smoothness derived from the models can distinguish whether a node pair belongs to different labels, we sample a fixed list of $n$ nodes from the last hidden representation derived from Cora and Pubmed. We divide the list of nodes into train/ validation/ test split following 7000 nodes in the train split, 2000 nodes in the validation split and 1000 nodes in the test split. Then we use a linear classifier to judge whether every node pair belongs to the same label. To measure node representations against over-smoothness of different methods, we use the AUC as an evaluating metric in Table \ref{tb:sample}. We can conclude that GraTO outperforms other baselines on Cora and Pubmed. This indicates that our node representation against over-smoothness can distinguish different-label nodes clearly which further proves our strong node representation against over-smoothness.

\begin{table}[t]
  \caption{Effectiveness of Drop-Attr. GraTO* refers to the model after replacing Drop-Attr.}
  \label{tb:dropattr}
  \scalebox{0.9}{
  \begin{tabular}{l|cc|cc|cc}
  \toprule
  \multirow{2}*{\diagbox{Method}{Dataset}} & \multicolumn{2}{c|}{Cora} & 
  \multicolumn{2}{c|}{Citeseer}& \multicolumn{2}{c}{Pubmed}\\
  & F1 & MAD & F1 & MAD & F1 & MAD\\
  \midrule
    GraTO*  
        & 77.0 & 82.2
        & 55.2 & 84.2 
        & \textbf{77.7} & 82.5\\
    GraTO 
        &\textbf{79.3} &\textbf{84.9} 
        &\textbf{61.4} & \textbf{84.5}
        & \textbf{77.7} & \textbf{87.5}\\
  \bottomrule
  \end{tabular}}
\end{table}

\subsection{Model Framework Study}\label{task:ablation}

\subsubsection{Effectiveness of Drop-Attr}
To prove the effectiveness of Drop-Attr, we replace the Drop-Attr operations as Skip-Connect, which only transfers $X$ and $A$ without any changes, in the block. 
The result is shown in Table \ref{tb:dropattr} and we can demonstrate that GraTO with Drop-Attrs outperforms the one without them. In Cora, Citeseer and Pubmed, Drop-Attr promotes the F1-score and MAD value of GraTO obviously. From the result above, we can demonstrate that Drop-Attr has a significant effect on promoting the model performance which further implies changing several parts of the nodes’ values on the feature can promote the model performance and alleviate the over-smoothing problem.
\subsubsection{Block Settings Study}
As we set the intermediate nodes as four in the previous study, we change the number of nodes from three to five. Due to the space limit, we don't set the number of intermediate nodes like 6 or above. After setting the number, we search for a block architecture from scratch following Algorithm \ref{algo}. The result is shown in Table \ref{tb:ablation} which shows that three-node block and five-node block outperform GraTO in MAD values but get a poor F1 score. Considering the model performance, the set of four nodes in a block is the best one.

\begin{table}
  \caption{Block Setting Study. The GraTO-n denotes that there are $n$ intermediate nodes in a block.}
  \label{tb:ablation}
  \begin{tabular}{l|cc|cc}
  \toprule
  \multirow{2}*{\diagbox{Method}{Dataset}} & \multicolumn{2}{c|}{Cora} & 
  \multicolumn{2}{c}{Citeseer}\\
  & F1 & MAD & F1 & MAD \\ 
  
  \midrule
    GraTO-3  
        & 67.9 & \textbf{91.4}
        & 48.5 &  \textbf{89.5}  \\
    \textbf{GraTO-4} 
        &\textbf{80.2
        } &  85.7 
        &\textbf{61.3} & 84.5\\
    GraTO-5
        &65.8 &  90.9 
        &53.2 & 89.3\\
    
  \bottomrule
  \end{tabular}
\end{table}

\section{Conclusion}\label{conclusion}
In this paper, we propose a framework that can automatically search for GNN architecture to tackle the over-smoothing problem. GraTO outperforms other models in MAD values and achieves competitive results in accuracy on node classification task and model's robustness task. We prove that the effectiveness of the loss function we propose in striking a balance between model performance and representation smoothness in the related experiments. We add several existing tackle over-smoothing methods and Drop-Attr which is proved to be effective in the NAS search space to jointly leverages multiple solutions to the problem. 
We will further explore a more flexible model structure in the future.

\section*{Acknowledgements}
This work was supported by the National Key Research and Development Program of China (No. 2020AAA0108800), National Nature Science Foundation of China (No. 62192781, No. 61872287, No. 61937001, No. 62250009, No. 62137002), Innovative Research Group of the National Natural Science Foundation of China (61721002), Innovation Research Team of Ministry of Education (IRT\_17R86), Project of China Knowledge Center for Engineering Science and Technology, Project of Chinese academy of engineering ``The Online and Offline Mixed Educational Service System for ‘The Belt and Road’ Training in MOOC China'' and CCF-AFSG Research Fund.

We would like to thank the reviewers and area chair for their constructive feedback. We would also like to thank all LUD lab members for our collaborative research environment. Shangbin Feng did this work while attending Xi'an Jiaotong University.

\bibliographystyle{ACM-Reference-Format}
\balance
\bibliography{mybib}

\end{document}